%% file: main.tex
\documentclass[10pt,twocolumn,letterpaper]{article}

\usepackage{cvpr}
\usepackage{times}
\usepackage{epsfig}
\usepackage{graphicx}
\usepackage{amsmath}
\usepackage{amssymb}
\usepackage{bm}
\usepackage{algorithm}
\usepackage{algpseudocode}
\usepackage{mathtools} %
\usepackage{dsfont} %

\newcommand{\vect}[1]{\boldsymbol{#1}}
\newcommand{\module}[1]{\textsc{#1}}
\newcommand{\func}[1]{\textsc{#1}}
\DeclareMathOperator*{\argmax}{arg\,max}
\DeclareMathOperator*{\softmax}{softmax}
\DeclareMathOperator*{\lnorm}{L2Norm}
\DeclareMathOperator*{\mlp}{MLP}
\DeclareMathOperator*{\onehot}{one\_hot}
\DeclareMathOperator*{\GSSampler}{GS\_Sampler}

\usepackage[pagebackref=true,breaklinks=true,letterpaper=true,colorlinks,bookmarks=false]{hyperref}

\cvprfinalcopy %

\def\cvprPaperID{3129} %

\ifcvprfinal\fi
\begin{document}

\title{Recursive Visual Attention in Visual Dialog}

\author{Yulei Niu$^{1,2}$~~~Hanwang Zhang$^{2}$~~~Manli Zhang$^{1}$~~~Jianhong Zhang$^{1}$~~~Zhiwu Lu$^{1}$\thanks{Corresponding author.}~~~~Ji-Rong Wen$^{1}$\\
$^{1}$Beijing Key Laboratory of Big Data Management and Analysis Methods\\
School of Information, Renmin University of China, Beijing 100872, China\\
$^{2}$Nanyang Technological University, Singapore 639798\\
{\tt\small \{niu, manlizhang, jianhong, luzhiwu, jrwen\}@ruc.edu.cn, hanwangzhang@ntu.edu.sg}
}

\maketitle

\input{section/0-Abstract}

\input{section/1-Introduction}
\input{section/2-Related_Work}
\input{section/3-Approach}
\input{section/4-Experiments}
\input{section/5-Conclusion}
\input{section/6-Acknowledgements}

{
\bibliographystyle{ieee}
\bibliography{main}
}

\end{document}

%% file: section/0-Abstract.tex
\begin{abstract}
Visual dialog is a challenging vision-language task, which requires the agent to answer multi-round questions about an image. It typically needs to address two major problems: (1) How to answer visually-grounded questions, which is the core challenge in visual question answering (VQA); (2) How to infer the co-reference between questions and the dialog history. An example of visual co-reference is: pronouns (\eg, ``they'') in the question (\eg, ``Are they on or off?'') are linked with nouns (\eg, ``lamps'') appearing in the dialog history (\eg, ``How many lamps are there?'') and the object grounded in the image. In this work, to resolve the visual co-reference for visual dialog, we propose a novel attention mechanism called Recursive Visual Attention (RvA). Specifically, our dialog agent browses the dialog history until the agent has sufficient confidence in the visual co-reference resolution, and refines the visual attention recursively. The quantitative and qualitative experimental results on the large-scale VisDial v0.9 and v1.0 datasets demonstrate that the proposed RvA not only outperforms the state-of-the-art methods, but also achieves reasonable recursion and interpretable attention maps without additional annotations. The code is available at \url{https://github.com/yuleiniu/rva}.
\end{abstract}
\vspace{-3mm}

%% file: section/1-Introduction.tex
\section{Introduction}

Vision and language understanding has become an attractive and challenging interdisciplinary field in computer vision and natural language processing. Thanks to the rapid development of deep neural networks and the high quality of large-scale real-world datasets, researchers have achieved inspiring progress in a range of vision-language tasks, including visual relation detection~\cite{lu2016visual,krishna2017visual,zhang2017ppr}, image captioning~\cite{vinyals2015show,chen2017sca,xu2015show,anderson2018bottom}, referring expression grounding~\cite{mao2016generation,nagaraja2016modeling,zhang2018grounding}, and visual question answering (VQA)~\cite{antol2015vqa,tapaswi2016movieqa,fukui2016multimodal,tang2018learning}. However, comprehension and reasoning in vision and natural language are still far from being resolved, especially when the AI agent interacts with human in a continuous communication, such as vision-and-language navigation~\cite{anderson2018vision} and visual dialog~\cite{das2017visual}.

\begin{figure}
    \centering
    \includegraphics[width=.96\linewidth]{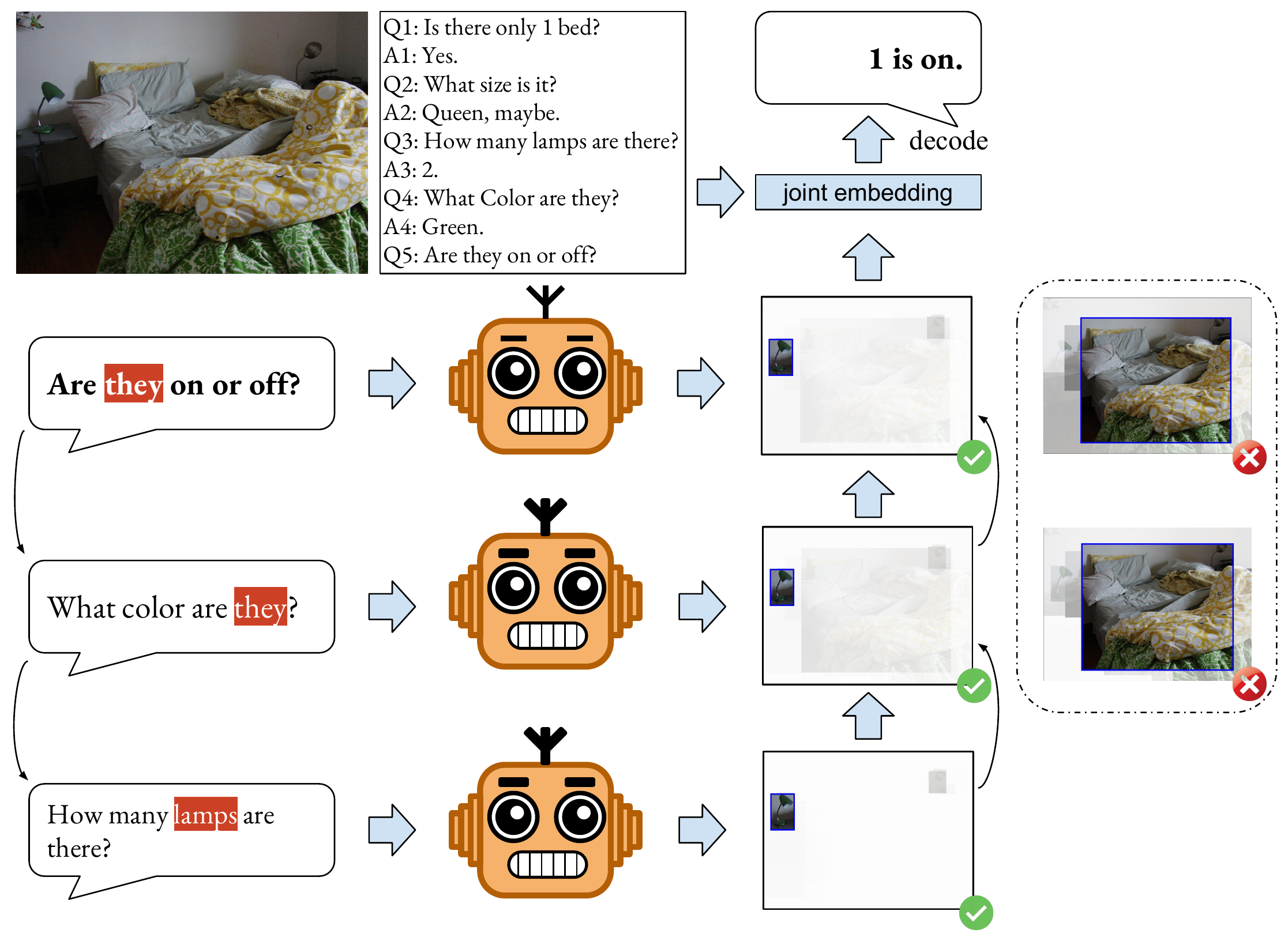}
    \caption{Illustration of the intuition of Recursive Visual Attention in visual dialog. When our dialog agent meets an ambiguous question (\eg, ``Are \underline{they} on or off?''), it will recursively review the dialog history (see the first column) and refine the visual attention (see the third column), until it can resolve the visual co-reference (\eg, How many \underline{lamps} are there?). The attention maps tagged with green check mark represent reasonable recursive visual attention, while those tagged with red cross mark in the dashed box represent false question-guided visual attention.}
    \label{fig:dessert}
    \vspace{-5mm}
\end{figure}

Visual dialog is one of the prototype tasks introduced in recent years~\cite{das2017visual,de2017guesswhat}. It can be viewed as the generalization of VQA, which requires the agent to answer the question about an image~\cite{antol2015vqa} or video~\cite{tapaswi2016movieqa} after comprehending and reasoning out of visual and textual contents. Different from one-round VQA, visual dialog is a multi-round conversation about an image. Therefore, one of the key challenges in visual dialog is visual co-reference resolution, since $98\%$ of dialogs and $38\%$ of questions in the large-scale VisDial dataset have at least one pronoun (\eg, ``it'', ``they'', ``this'', ``he'', ``she'')~\cite{das2017visual}. For example, as illustrated in Figure~\ref{fig:dessert}, questions ``Are \underline{they} on or off?'' and ``What color is \underline{it}?'' contain pronouns that need to be resolved before answering. Recently, researchers have attempted to resolve the visual co-reference using attention memory~\cite{seo2017visual} at a sentence level, or applying the neural module networks~\cite{kottur2018visual} at a word level. Specifically, an attention memory~\cite{seo2017visual} is established to store the image attention map at each round, while a reference pool~\cite{kottur2018visual} is utilized to keep all the entities recognized from the dialog history. They both apply a soft attention over all the stored visual attentions for refinement. However, humans rarely remember all their previous visual attentions, and only review the topic-related dialog history when they are confused with the ambiguous question. 

We expect our dialog agent to \emph{selectively} review the dialog history like us humans during the conversation. For example, as illustrated in Figure~\ref{fig:dessert}, ``Are they on or off?'' is an ambiguous question and the dialog agent needs to resolve ``they'' before watching the image. The agent then \emph{recursively} browses the dialog history and computes visual attention until it meets the unambiguous description ``How many lamps are there?''. One may argue that a natural language parser can achieve this goal by detecting whether there exists a pronoun in the question. However, not all pronouns are needed to be resolved, \eg, ``Is it sunny?''. Some abbreviate sentences without context are also ambiguous, \eg, ``What color?''. It is thus impractical to exhaust all cases using a natural language parser.

In this work, we formulate visual co-reference resolution in visual dialog as Recursive Visual Attention (RvA). As shown in Figure~\ref{fig:dessert}, the agent first infers whether it can ground the visual content based on the current question. If not, the agent will recursively review the topic-related dialog history and refine the visual attention. The recursion termination is that the agent feels ``confident'' in visual grounding, or it has backtracked to the beginning of dialog history. Thanks to the Gumbel-Max trick~\cite{gumbel1954statistical} and its continuous softmax relaxation~\cite{jang2016categorical,maddison2016concrete}, our agent can be end-to-end trained when making discrete decisions. In addition, we design two types of language features for different purposes. The \textit{reference-aware} language feature helps with visual grounding and inference of reviewing dialog history, while the \textit{answering-aware} language feature controls which attributes of the image feature should be activated for question answering.

Our main contributions are concluded as follows. First, we propose a novel Recursive Visual Attention (RvA) strategy for the visual co-reference resolution in visual dialog. Second, we carry out extensive experiments on VisDial v0.9 and v1.0~\cite{das2017visual}, and achieve state-of-the-art performances compared to other methods. Third, the qualitative results indicate that our dialog agent obtains reliable visual and language attention during the reasonable and history-aware recursive process.

%% file: section/2-Related_Work.tex
\section{Related Work}

\noindent \textbf{Visual Dialog.} 
Visual dialog is a current vision and language task, which requires the agent to understand the dialog history, ground visual object, and answer the question. Recently, two popular dialog datasets were crowd-sourced on Amazon Mechanical Turk (AMT) \cite{buhrmester2011amazon}. De Vries \etal \cite{de2017guesswhat} collected GuessWhat dataset from a cooperative two-player game. Given the whole picture and its caption, one player asks questions to locate the selected object, while the other player replies in yes/no/NA. However, the questions are constrained to closed-ended questions. In comparison, Das \etal \cite{das2017visual} collected VisDial dataset by a different two-person chat style. During the live chat, the ``questioner'' asks questions to imagine the visual content in the picture based on the caption and chat history, while the ``answerer'' watches the picture and answer in a free-form way. We apply the second setting in this paper. 

\noindent \textbf{Visual Co-reference Resolution.} The task of visual co-reference resolution is to link expressions, typically pronoun and noun phrases referring to the same entity, and ground the referent in the visual content. Co-reference resolution has been used to improve visual comprehension in many tasks, such as visual grounding~\cite{huang2018finding}, action recognition~\cite{ramanathan2014linking,rohrbach2017generating}, and scene understanding~\cite{kong2014you}. Recently Lu \etal~\cite{lu2017best} proposed a history-conditioned attention mechanism to implicitly resolve the visual co-reference. Seo \etal~\cite{seo2017visual} used attention memory to store previous image attentions at a sentence level. Furthermore, neural module networks~\cite{andreas2016neural} were applied to recognize entities in all the history at a word level~\cite{kottur2018visual}. Different from recent works that proposed a soft attention mechanism over all the memorized attention maps~\cite{seo2017visual} or all the grounded entities~\cite{kottur2018visual}, our proposed recursion predicts \textit{discrete} attention over topic-related history, which is more intuitive and explainable.

%% file: section/3-Approach.tex
\vspace{-3mm}
\section{Approach}\label{sec:approach}
\vspace{-1mm}
In this section, we formally introduce the visual dialog task and our proposed Recursive Visual Attention (RvA) approach. The task of visual dialog~\cite{das2017visual} is defined as follows. The dialog agent is expected to answer the question $q_T$ at round $T$ by ranking a list of 100 candidate answers $A_T=\{a_T^{(1)},\cdots,a_T^{(100)}\}$ in a discriminative manner, or producing a sentence in a generative manner. The extra information for visual dialog consists of the image $I$ and the dialog history $H=\{\underbrace{c}_{h_0}, \underbrace{(q_1,a_1)}_{h_1},\cdots,\underbrace{(q_{T-1},a_{T-1})}_{h_{T-1}}\}$, where $c$ is the image caption and $(q,a)$ is any question-answer pair.

Next, we first provide an overall structure of RvA in Section~\ref{sec:rva}, followed by Section~\ref{sec:modules} introducing the \module{Infer}, \module{Pair} and \module{Att} modules of RvA. The training details of RvA are given in Section~\ref{sec:details}.

\begin{figure}
    \centering
    \includegraphics[width=1\linewidth]{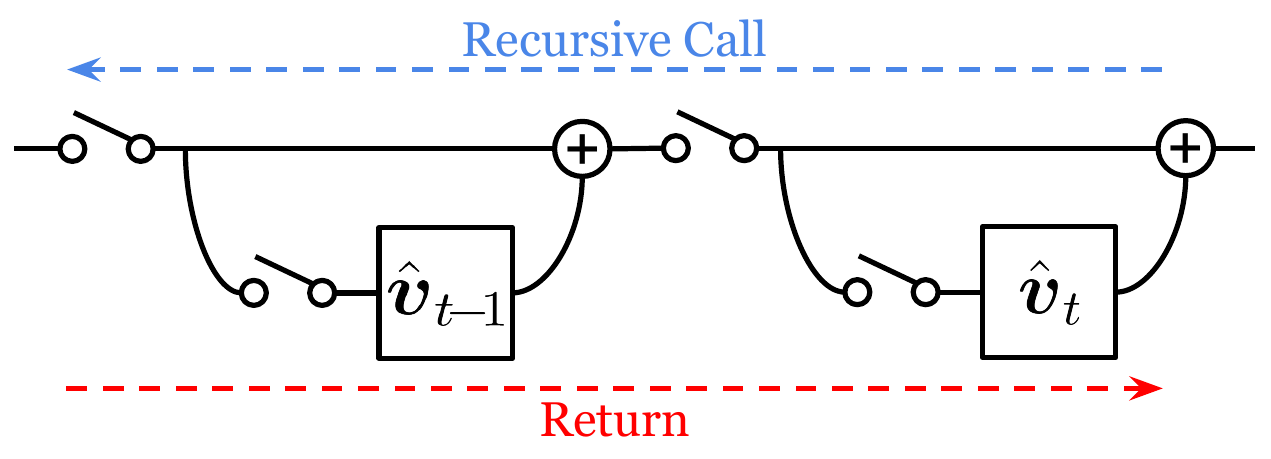}
    \caption{A high-level view of Recursive Visual Attention. The right-to-left direction (dashed blue) represents recursive call, and the left-to-right direction (dashed red) represents visual attention return. The $cond$ variable controls the switch on the trunk, while $t_p$ controls the switch on the branch (see Algorithm~\ref{alg:rva}). $\hat{\vect{v}}_t$ represents the attended feature $\func{Att}(\mathcal{V},\mathcal{Q},t)$.}
    \label{fig:rva}
    \vspace{-3mm}
\end{figure}

\subsection{Recursive Visual Attention}\label{sec:rva}

\input{alg/alg-rva}
\vspace{-1mm} 
First of all, the overall structure of the proposed Recursive Visual Attention (RvA) method is shown in Algorithm~\ref{alg:rva}. Here $\mathcal{Q}=\{\vect{q}_0,\vect{q}_1,\cdots,\vect{q}_T\}$ represents the question feature set where the caption feature $\vect{c}$ is added into the question set as $\vect{q}_0$, $\mathcal{H}=\{\vect{h}_0,\vect{h}_1,\cdots,\vect{h}_{T-1}\}$ represents the history feature set, and $\mathcal{V}=\{\vect{v}_1,\cdots,\vect{v}_K\}$ represents the region feature set. Given any question $q_t$, our dialog agent first \textit{infers} whether it understands the question $q_t$ for visual grounding. If not, our agent will \textit{pair} the current question $q_t$ with its most related history $h_{t_p}$, and backtrack to the paired round $t_p$. This process will be kept executing until the agent can understand the current traced question, or the dialog agent has backtracked to the beginning of the dialog. As a result, our dialog agent recursively modifies the visual attention by adding the question-guided \textit{attended} visual attention at round $t$ and the recursive visual attention at paired round $t_p$, weighted by a learnable non-negative weight $\lambda$. For the question $q_T$, the output visual attention is formulated by $\vect{\alpha}_T\!=\!\module{RvA}(\mathcal{V},\mathcal{Q},\mathcal{H},T)$. The attended visual feature is further calculated by a weighted sum over all the region features $\hat{\vect{v}}_T\!=\!\sum_{i}\alpha_i\vect{v}_i$.

In addition, we give a high-level view of Recursive Visual Attention (RvA) in Figure~\ref{fig:rva}. Intuitively, all the switches on both the trunk and branches are initially open (\ie, turned off). Our RvA is recursively called from \emph{present} to \emph{past}, closing (\ie, turning on) the switch on the trunk until the recursion terminates. The switch of the question-guided visual feature $\vect{v}_{t_p}$ on the branch is closed if the history $h_{t_p}$ is paired with the current traced question $q_t$. When the recursion termination condition is met, we unroll the process from past to present and finally obtain the recursive visual feature.

We further design three modules to achieve the recursive visual attention algorithm, \ie, \module{Infer} , \module{Pair}, and \module{Att} (\ie, attend). In overview, \module{Infer} module asserts the recursion termination condition and computes visual feature fusion weight, \module{Pair} module returns the paired round, and \module{Att} module calculates question-guided visual attention.

\subsection{Neural Modules}\label{sec:modules}
\vspace{-1mm} 
\input{alg/alg-infer}
\vspace{-2mm} 
\noindent \textbf{\module{Infer} Module.} \module{Infer} module is designed to 1) determine whether to review the dialog history, 2) provide a weight to fuse the recursive visual attention and the question-guided visual attention. Specifically, \module{Infer} module takes the question feature $\vect{q}_t$ as input. The outputs include 1) a Boolean $cond$ to decide whether to terminate the recursion, and 2) a weight $\lambda\!\in\!(0,1)$ for visual attention fusion.

\begin{figure}
    \centering
    \includegraphics[width=.98\linewidth]{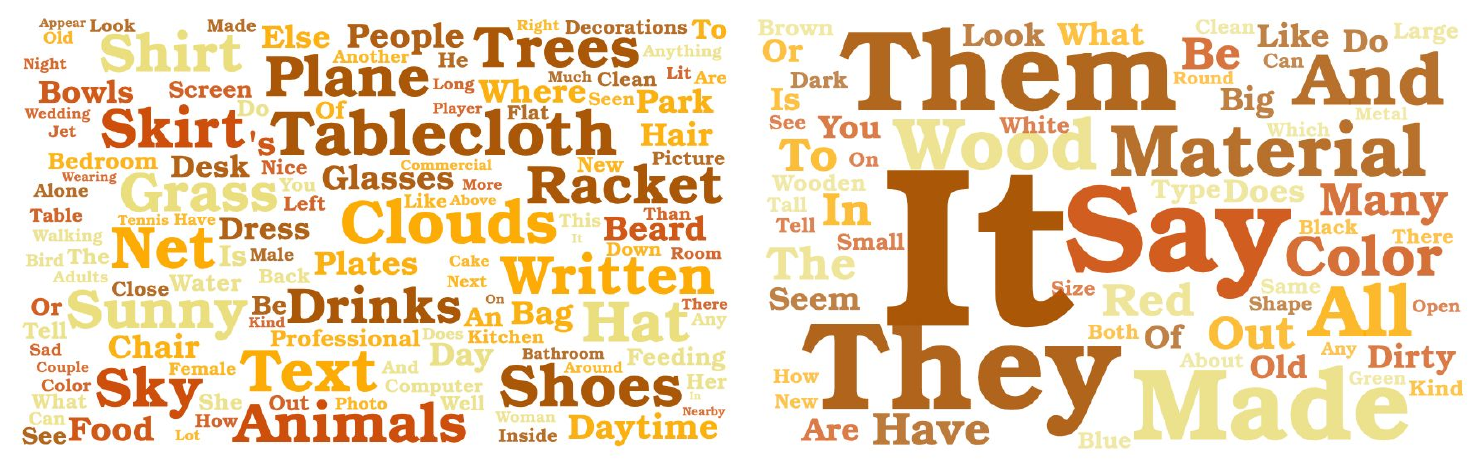}
    \caption{Word cloud visualization of word attention in RvA. For questions that our dialog agent thinks to be unambiguous (left), the word attentions are spread out a variety of nouns (\eg, ``clouds'', ``drinks''). For questions that confuse the agent (right), the word attention significantly focuses on pronouns (\eg, ``it'', ``they'').}
    \label{fig:wordcloud}
\end{figure}

The recursion will be terminated if at least one of the following conditions is satisfied (see lines 5-7 in Algorithm~\ref{alg:infer}). First, the review backtracks to the very starting point: caption. Second, the question $q_t$ is predicted to be unambiguous. In order to estimate the ambiguity of the question, we use a non-linear transformation~\cite{teney2017tips} $f^{I}_{q}(\cdot)$, followed by a Gumbel Sampling operation $\GSSampler$ for differentiable discrete decision:
\begin{align}
    \vect{z}^I_t &= f^{I}_{q}(\vect{q}_t);\\
    \quad\vect{o}^{I}_t &= \GSSampler(W^I\vect{z}^{I}_t)
\end{align}
where $W^{I}$ denotes the learnable parameters. $\GSSampler$ (see Section~\ref{sec:gumbel}) outputs a 2-dim one-hot vector $\vect{o}^{I}_{t}$ for discrete decision, where the binary element $o^I_{t,0}$ is encoded as the Boolean output to determine whether $q_t$ is ambiguous. As illustrated in Figure~\ref{fig:wordcloud}, our dialog agent successfully learns the relation between words and recursion termination without additional annotations.

\vspace{-5mm}

\input{alg/alg-pair}
\noindent \textbf{\module{Pair} Module.} We observe that an ambiguous question often follows the latest topic. A simple idea is to directly pair the question with its latest history, \ie, set $t_p$ as $t\!-\!1$ in \module{Infer} module. However, the questioner sometimes traces back to an earlier topic, which means that the question has no relationship with its latest history. Therefore, we design a \module{Pair} module to estimate which history is most related with the question $q_t$.

Algorithm~\ref{alg:pair} shows the structure of \module{Pair} module. Specifically, \module{Pair} module takes the question feature $\vect{q}_t$ and the history feature $\mathcal{H}=\{\vect{h}_0,\cdots,\vect{h}_{t\!-\!1}\}$ as input, and predicts which history is most related to $q_t$. The \module{Pair} module is formulated as:
\vspace{-2mm}
\begin{align}
    z^{P}_{t,i} &= \mlp([f^{P}_q(\vect{q}_t), f^{P}_h(\vect{h}_i)])\label{eq:2} \\
    \vect{o}^P_t &= \GSSampler(W^P[\vect{z}^P_t, \vect{\Delta}_{t}])\\
    t_p &= \sum^{t-1}_{i=0}o^P_{t,i}\cdot{i}
\end{align}

\noindent where $[\cdot]$ is the concatenation operation. The \module{Pair} module considers 1) the matching score between the question $q_t$ and the history $h_i$, which is denoted as $z^{P}_{t,i}$; 2) the ``sequential distance'' between $q_t$ and $h_i$ in the dialog, which is measured by $\Delta_{t,i}=t\!-\!i$. Finally, $\GSSampler$ outputs a $t$-dim one-hot vector $\vect{o}^P_t$ for discrete decision (\ie, pairing the question with a single history). The question $q_t$ will be paired with the $k$-th history $h_k$ if $o^P_{t,k}\!=\!1$, \ie, the $k$-th history $h_{k}$ matches the question $q_t$ better than others.

\input{alg/alg-att}
\noindent \textbf{\module{Att} Module.} \module{Att} module takes visual features of regions $V\!=\!\{\vect{v}_1,\cdots,\vect{v}_K\}$ and the question feature $\vect{q}_t$ as input, and outputs question-guided visual attention. As shown in Algorithm~\ref{alg:att}, the question-guided visual attention is formulated as:
\begin{align}
    \vect{z}^A_{t,i} &= \lnorm(f^A_q(\vect{q}_t)\circ f^A_v(\vect{v}_i)) \\
    \vect{\alpha}^A_t &= \softmax(W^A {\bm Z}^A_t)\label{eq:imgatt}
\end{align}
where $f^A_q(\cdot)$ and $f^A_v(\cdot)$ represents non-linear trans- formations to embed visual and language features into the same space, and $\circ$ denotes Hadamard (element-wise) product for multi-modal feature fusion.

\subsection{Training}\label{sec:details}

As mentioned in Section~\ref{sec:modules}, our Recursive Visual Attention takes visual and language representations as input, and applies Gumbel sampling for differentiable discrete decision. The details are given as follows.

\subsubsection{Feature Representation}\label{sec:feature}

\noindent\textbf{Language Feature.} Let $\mathcal{W}^q_t=\{\vect{w}^q_{t,1},\cdots,\vect{w}^q_{t,m}\}$ be the word embeddings of the question $q_t$. The word embeddings are passed through the bidirectional LSTM (bi-LSTM):
\begin{align}
    \overrightarrow{\vect{h}}^q_{t,i} &= \mathrm{LSTM}^q_f(\vect{w}^q_{t,i}, \overrightarrow{\vect{h}}^q_{t,i-1}) \\
    \overleftarrow{\vect{h}}^q_{t,i} &= \mathrm{LSTM}^q_b(\vect{w}^q_{t,i}, \overleftarrow{\vect{h}}^q_{t,i+1})\\
    \vect{h}^q_{t,i} &=[\overrightarrow{\vect{h}}^q_{t,i},\overleftarrow{\vect{h}}^q_{t,i}]
\end{align}
\noindent where $\overrightarrow{\vect{h}}^q_{t,i}$ and $\overleftarrow{\vect{h}}^q_{t,i}$ represent forward and backward hidden state of the $i$-th word respectively, $\mathrm{LSTM}^q_f$ and $\mathrm{LSTM}^q_b$ represent the forward and backward LSTMs. We use the concatenation of last hidden states $\vect{e}^q_t\!=\![\overrightarrow{\vect{h}}^q_{t,m},\overleftarrow{\vect{h}}^q_{t,1}]$ as the encoding of the whole question $q_t$. Similarly, we can encode the history $h_i$ as $\vect{e}^h_i$ using the same bi-LSTM with different parameters. In \module{Pair} module, we denote $\vect{e}^q_t$ as $\vect{q}_t$ and $\vect{e}^h_i$ as $\vect{h}_i$ to calculate the matching score between the question $q_t$ and the history $h_i$.

\begin{figure} 
\centering 
    \includegraphics[width=.9\linewidth]{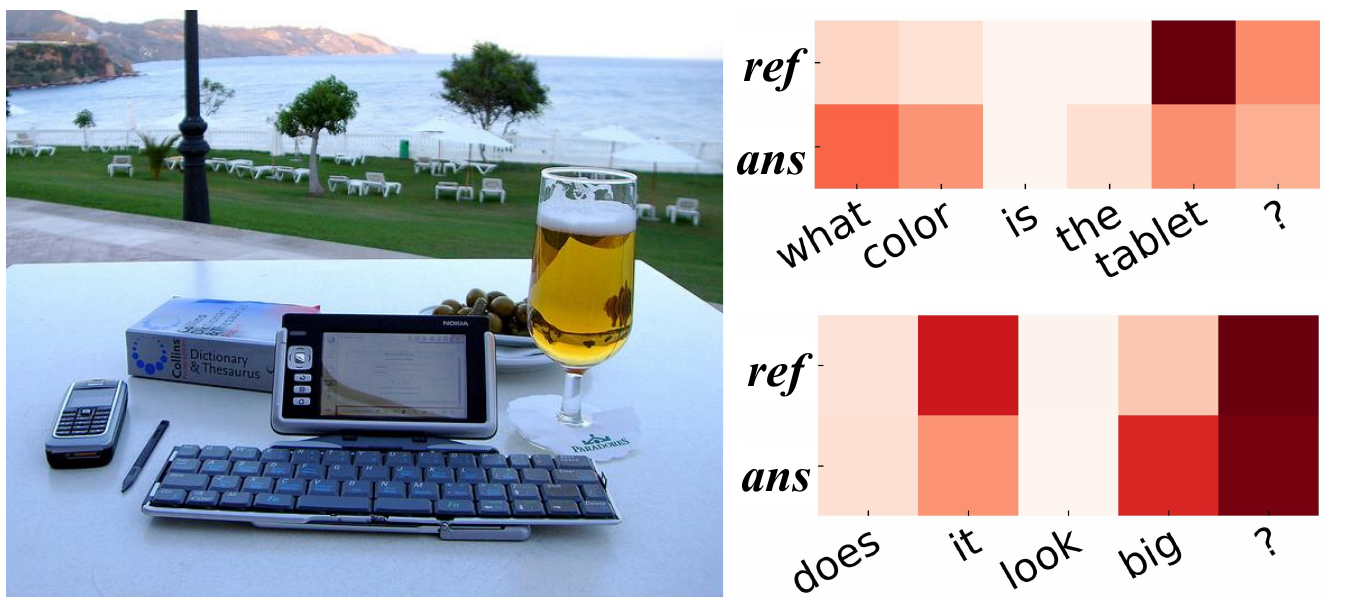} \caption{A qualitative example of question attentions. The reference-aware (\textit{ref}) question attention mainly emphasizes nouns (\ie, \textit{``tablet''}) and pronouns (\ie, \textit{``it''}) for recursion termination estimation and visual grounding. The answering-aware (\textit{ans}) question attention highlights property words (\ie, \textit{``what color'', ``big''}) to record question type and activate specific attributes of visual representation for question answering. Darker color indicates higher weight.} \label{fig:langfeat}
    \vspace{-3mm} 
\end{figure}

Note that the words contribute differently to the question representation for various purposes. An example is illustrated in Figure~\ref{fig:langfeat}. On one hand, the words ``tablet'' and ``it'' should be emphasized for recursion termination estimation and visual grounding. On the other hand, the phrase ``what color'' and the word ``big'' should be highlighted to activate specific attributes of the visual representation for question answering. Therefore, we encode each question using self-attention mechanisms~\cite{vaswani2017attention} into two forms: \textit{reference-aware} question feature $\vect{q}^{ref}_t$  and \textit{answering-aware} question feature $\vect{q}^{ans}_t$. Different from prior attention mechanism that uses linear transformation followed by hyperbolic tangent ($\tanh$) activation, we formulate the self-attention mechanism as:
\begin{align}
    \vect{z}^{q,*}_{t,i} &= \lnorm(f^{q,*}_q(\vect{h}^q_{t,i})) \\
    \vect{\alpha}^{q,*}_t &=\softmax(W^{q,*}{\bm Z}^{q,*}_t) \\
    \vect{q}^*_t &= \sum^m_{i=1}\alpha^{q,*}_{t,i}\vect{w}^q_{i}
\end{align}
where $f^{q,*}_q(\cdot)$ is a non-linear transformation function, $W^{q,*}$ is the learnable parameters, and $*\!\in\!\{ref,ans\}$. The attended question features $\vect{q}^{ref}_t$ and $\vect{q}^{ans}_t$ are calculated by a weighted sum overall the words. In \module{Infer} and \module{Att} modules, we denote $\vect{q}^{ref}_t$ as $\vect{q}_t$ for recursion termination estimation and visual grounding.

\noindent\textbf{Visual Feature.} 
Spatial image features with attention mechanism have been widely used in many vision and language tasks, such as image captioning and visual question answering. Recently, a bottom-up attention mechanism~\cite{anderson2018bottom} is proposed based on the Faster R-CNN framework. The ResNet model is utilized as backbone and trained on Visual Genome~\cite{krishna2017visual} dataset to predict attributes and classes. In this paper, we apply the bottom-up attention mechanism and select top-$K$ region proposals from each image, where $K$ is simply fixed as $36$.

After obtaining the visual feature $\hat{\vect{v}}_T$ using Recursive Visual Attention, we further refine the visual feature using the answering-aware question feature $\vect{q}^{ans}_T$. The motivation is that only question-related attributes of visual content are useful for answering questions (\eg, ``\underline{What} \underline{color} is the tablet?'', ``Does it look \underline{big}?'' in Figure~\ref{fig:langfeat}). Motivated by the gating operations within LSTMs and GRUs, we further refine visual feature as:
\begin{equation}
    \tilde{\vect{v}}_T =\hat{\vect{v}}_T\circ f^v_q(\vect{q}^{ans}_T)
    \label{eq:filter}
\end{equation}
where the output of non-linear transformation $f^v_q(\cdot)$ works as a ``visual feature filter'' to deactivate the information unrelated to answering questions in the visual representation $\hat{\vect{v}}_t$.

\noindent\textbf{Joint Embedding.} Considering that the dialog history reflects prior knowledge of visual content, we obtain the ``fact'' embedding by attending to all the history as
\begin{align}
    \vect{z}^h_{T,i} &= \lnorm(f^h_q(\vect{e}^q_T)\circ f^h_h(\vect{e}^h_i)) \\ \label{eq:factatt}
    \vect{\alpha}^h_T &=\softmax(W^h{\bm Z}^h_{T})\\
    \vect{h}^{f}_T & =\sum^{T-1}_{i=0}\alpha^h_{T,i}\vect{e}^h_{i}
\end{align}
where $f^h_h$ and $f^h_q$ are non-linear transformation functions. The ``fact'' embedding $\vect{h}^{f}_T$ is calculated by a weighted sum over all the history encodings.

Since we have obtained the filtered visual feature $\tilde{\vect{v}}_T$, the answering-aware question feature $\vect{q}^{ans}_T$, and the fact embedding $\vect{h}^f_T$ for question $q_T$, we concatenate these features and use a linear transform followed by a tangent activation to obtain the final joint embedding:
\begin{equation}
\vect{e}^J_T=\tanh(W^J[\tilde{\vect{v}}_T, \vect{q}^{ans}_T, \vect{h}^{f}_T])
\end{equation}
where $[\cdot]$ denotes the concatenation operation. The joint embedding is further fed into the answering decoder.

\subsubsection{Gumbel Sampling}\label{sec:gumbel}

Our dialog agent needs to make a discrete decision in some cases, \eg, estimating whether to review the history and which history should be paired. In addition, we hope that the gradients can be back propagated through discrete decision making for end-to-end training. In order to achieve these goals, we utilize the Gumbel-Max trick~\cite{gumbel1954statistical} with its continuous softmax relaxation~\cite{jang2016categorical,maddison2016concrete}. Specifically, the samples $\vect{z}$ can be drawn from a categorical distribution with $\vect{\pi}\!=\!\{\pi_1,\cdots,\pi_c\}$ as:
\begin{equation}\label{eq:argmax}
    \vect{z} = \onehot\left(\argmax_{k\in\{1,\dots,c\}}(\log(\pi_k)+g_k)\right)
\end{equation}
where $\vect{g}\!=\!-\!\log(\!-\!\log(\vect{u}))$ with $\vect{u}\!\sim\!\text{unif}[0,1]$.

The softmax relaxation of Gumbel-Max trick is to replace non-differentiable $\argmax$ operation with the continuous $\softmax$ function: 
\begin{equation}\label{eq:softmax}
\begin{split}
    \hat{\vect{z}} & = \softmax\left((\log(\vect{\pi})+\vect{g})/\tau\right) \\
\end{split}
\end{equation}
where the temperature of the softmax function $\tau$ is empirically set as $1$ in our work. During the training stage, we obtain an one-hot vector $\vect{z}$ as the discrete sample from Eq.~\ref{eq:argmax} for forward propagation, and compute gradients w.r.t. $\vect{\pi}$ in Eq.~\ref{eq:softmax} for back propagation. At the test stage, we greedily draw the sample with the largest probability without Gumbel samples $\vect{g}$.

%% file: alg/alg-rva.tex
\begin{algorithm}[h]
\caption{Recursive Visual Attention}\label{alg:rva}
\begin{algorithmic}[1]
\Function {RvA}{$\mathcal{V},\mathcal{Q},\mathcal{H}, t$}
    \State $cond, \lambda \leftarrow \func{Infer}(\mathcal{Q},t)$
    \If {$cond$} 
        \State \textbf{return} $\func{Att}(\mathcal{V},\mathcal{Q},t)$
    \Else
        \State $t_p \leftarrow\func{Pair}(\mathcal{Q},\mathcal{H},t)$
        \State \textbf{return} $(1\!-\!\lambda)\cdot\func{RvA}(\mathcal{V},\mathcal{Q},\mathcal{H},t_p)$
        \State $\qquad\quad+\lambda\cdot\func{Att}(\mathcal{V},\mathcal{Q},t)$
    \EndIf
\EndFunction
\end{algorithmic}
\end{algorithm}

%% file: alg/alg-infer.tex
\begin{algorithm}
\caption{\module{Infer} Module}\label{alg:infer}
  \begin{algorithmic}[1]
  \Function {Infer}{$\mathcal{Q},t$}
    \State $\vect{z}^I_t\leftarrow f^I_q(\vect{q}_t)$
    \State $\vect{o}^I_t\leftarrow\GSSampler(W^I\vect{z}_t^I)$
    \State $\vect{\alpha}^I_t\leftarrow\softmax(W^I\vect{z}_t^I)$
    \State $cond_1 \leftarrow t\!\stackrel{?}{=}\!0$
    \State $cond_2 \leftarrow o^I_{t,0}\!\stackrel{?}{=}\!1$
    \State $cond  \leftarrow cond_1\text{ or } cond_2$ \Comment{recursion termination}
    \State $\lambda \leftarrow \alpha^I_{t,0}$ \Comment{attention fusion weight}
    \State \textbf{return} $cond, \lambda$
    \EndFunction
\end{algorithmic}
\end{algorithm}

%% file: alg/alg-pair.tex
\begin{algorithm}
\caption{\module{Pair} Module}\label{alg:pair}
  \begin{algorithmic}[1]
  \Function {Pair}{$\mathcal{Q},\mathcal{H},t$}
      \State $\vect{e}^q_t \leftarrow f^P_q(\vect{q}_t)$
      \For {$i \leftarrow 0, \cdots, t\!-\!1$}
          \State $\vect{e}^h_i \leftarrow f^P_h(\vect{h}_i)$
          \State $z^P_{t,i} \leftarrow \mlp([\vect{e}^q_t,\vect{e}^h_i])$
          \State $\Delta_{t,i} \leftarrow t\!-\!i$
      \EndFor
      \State $\vect{o}^P_t \leftarrow \GSSampler(W^P[\vect{z}^P_t, \vect{\Delta}_{t}])$
      \State $t_p \leftarrow \sum_i o_{t,i}^P\cdot i$
      \State \textbf{return} $t_p$
  \EndFunction
\end{algorithmic}
\end{algorithm}

%% file: alg/alg-att.tex
\begin{algorithm}
\caption{\module{Att} Module}\label{alg:att}
    \begin{algorithmic}[1]
    \Function {Att}{$\mathcal{V},\mathcal{Q},t$}
        \State $\vect{e}^q_t\leftarrow f^A_q(\vect{q}_t)$
        \For {$i \leftarrow 1, \cdots, K$}
            \State $\vect{e}^v_i \leftarrow f^A_v(\vect{v}_i)$
            \State $\vect{z}^A_{t,i} \leftarrow \lnorm(\vect{e}^q_t\circ\vect{e}^v_i)$
        \EndFor
        \State $\vect{\alpha}^A_t \leftarrow \softmax(W^A{\bm Z}^A_{t})$
        \State \textbf{return} $\vect{\alpha}^A_t$
    \EndFunction
\end{algorithmic}
\end{algorithm}

%% file: section/4-Experiments.tex
\section{Experiments}

Our proposed model is evaluated on two real-world datasets: VisDial v0.9 and v1.0~\cite{das2017visual}. In this section, we first introduce the datasets, evaluation metrics, and implementation details. We then compare our method with the state-of-the-art models and provide qualitative results. 

\subsection{Datasets and Setup}
The VisDial v0.9~\cite{das2017visual} dataset was collected based on MS-COCO~\cite{lin2014microsoft} images and captions. In a two-player chat game, one player attempts to learn about an unseen image and asks questions based on the previous dialog, while the other player watches the image and replies with free-form answers. The whole chat lasts for 10 rounds for each image. As a result, the VisDial v0.9 dataset contains 83k dialogs on MS-COCO training images and 40k dialogs on validation images. Recently, the VisDial v1.0~\cite{das2017visual} dataset was released, including additional 10k dialogs on Flickr images. The collection of dialogs on Flickr images is similar to that on MS-COCO images. Overall, the new train split consists of 123k dialogs on MS-COCO images, which is the combination of train and validation splits from VisDial v0.9. The validation and test splits have 2k and 8k dialogs on Flickr images, respectively. Different from val split in VisDial v0.9 where each image is associated with a 10-round dialog, the dialogs in VisDial v1.0 test split have a random length within 10 rounds. 

\subsection{Metrics}

As in \cite{das2017visual}, we evaluated the responses at each round in VisDial v0.9 and the last round in VisDial v1.0 in a retrieval setting. Specifically, at test stage, each question is linked with a list of 100 candidate answers. The model is expected to rank over the candidates and return a ranked list for further evaluation. The metrics for retrieval performance evaluation are: 1) mean rank of human response (\textbf{Mean}); 2) recall@$k$ (\textbf{R@k}), which is the existence of the human response in the top-$k$ responses; 3) mean reciprocal rank (\textbf{MRR}) of the human response in the returned ranked list. As for VisDial v1.0, we also used the newly introduced normalized discounted cumulative gain (\textbf{NDCG}), which penalizes the lower rank of answers with high relevance. 

\subsection{Implementation Details}

\noindent \textbf{Language Model.} We pre-processed the text data as follows. As in \cite{das2017visual}, we first lowercased all questions and answers, converted digits to words, and removed contractions, before tokenizing using the Python NLTK toolkit~\cite{nltk}. The captions, questions, and answers were then padded or truncated to 40, 20 and 20, respectively. We kept words to those that occur at least 5 times in the training split, resulting in a vocabulary of 9,795 words for VisDial v0.9 and 11,336 words for VisDial v1.0. Our word embeddings are 300-dim vectors, initialized with pre-trained GloVe \cite{pennington2014glove} embeddings and shared across captions, questions and answers. The dimension of hidden states in all LSTMs is set to 512 in this work.

\noindent \textbf{Training Details.} We minimized the standard cross- entropy loss for the discriminative training, and the maximum likelihood estimation (MLE) loss for generative training. We used Adam \cite{kingma2014adam} with the learning rate of $1\!\times\!10^{-3}$, multiplied by 0.5 after every epoch, decreasing to $5\!\times\!10^{-5}$. We also applied Dropout~\cite{srivastava2014dropout} with a ratio of 0.5 before each fully-connected layer. Other settings are default in PyTorch~\cite{pytorch}.

\subsection{Comparing Methods}

We compared our proposed Recursive Visual Attention (\textbf{RvA}) model with the state-of-the-art methods in both discriminative and generative settings. Based on the design of encoders, these methods can be grouped into:

\noindent \textbf{Fusion-based Models.} Early methods simply fuse image, question, and history features at different stages. These early methods  include \textbf{LF}~\cite{das2017visual} and \textbf{HRE}~\cite{das2017visual}.

\begin{figure*}
    \centering
    \includegraphics[width=0.92\linewidth]{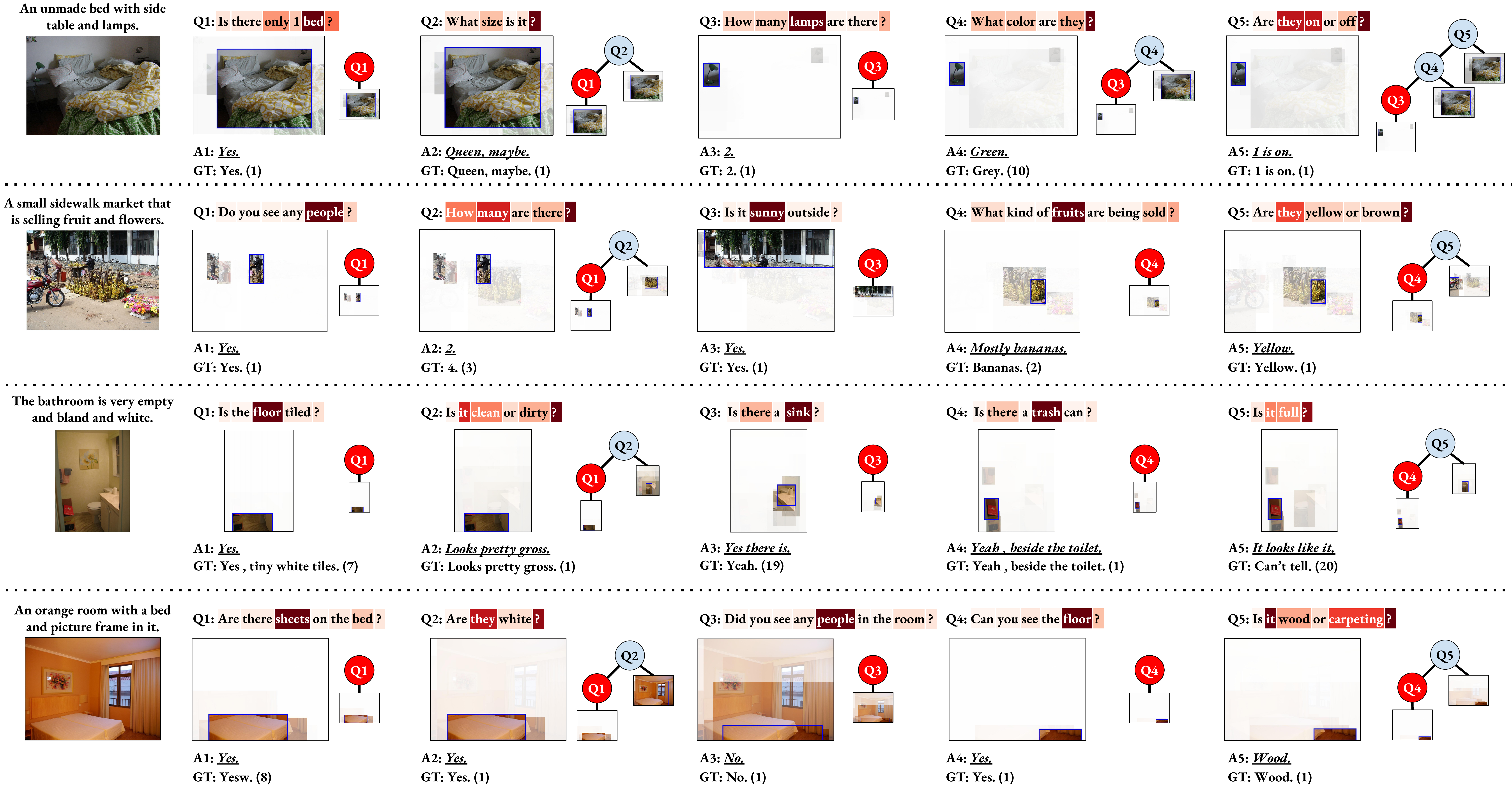}
    \caption{Qualitative results of our RvA model on VisDial dataset. The number in the bracket is the rank of ground-truth (GT) answer in the returned sorted list. Our model successfully obtains interpretable reference-aware question attention (represented by the highlight color of words, darker color indicates higher weight), reliable recursive image attention (represented by the attention map below the question), and reasonable recursions (represented by the recursion tree). The root nodes in the recursion tree represent the questions to be answered, the red nodes denote the questions terminating recursions, and the leaf nodes represent question-guided visual attention.}
    \label{fig:results}
    \vspace{-3mm}
\end{figure*}

\noindent \textbf{Attention-based Models.} Furthermore, some methods establish attention mechanisms over image, question, and history. The attention-based methods include \textbf{HREA}~\cite{das2017visual}, \textbf{MN}~\cite{das2017visual}, \textbf{HCIAE}~\cite{lu2017best}, and \textbf{CoAtt}~\cite{wu2017you}.

\noindent \textbf{VCoR-based models.} Recent works have focused on explicit visual co-reference resolution (VCoR) in visual dialog. We compared our method with VCoR-based models including \textbf{AMEM}~\cite{seo2017visual} and \textbf{CorefNMN}~\cite{kottur2018visual}.

\noindent \textbf{Ablative Models.} In addition, we evaluate the individual contribution of following features and components in our method: 1) \textbf{RPN}: we replaced the region proposal network with VGG-16~\cite{simonyan2014very} model, and used the spatial grids of \textit{pool5} feature map as regions. 2) \textbf{Bi-LSTM}: we replaced bidirectional LSTM with the vanilla LSTM. 3) \textbf{Rv}: we only considered the termination condition of RvA, and replaced the recursive attention with question-guided attention. 4) \textbf{FL}: we withdrew the ``visual feature filter'' $f^v_q(\cdot)$ in Eq.~\ref{eq:filter}, which controls the activation of visual attributes.

\input{tab/tab1-VisDial0.9-disc}

\subsection{Quantitative Results}

Table~\ref{tab-0.9-disc} reports the retrieval performances of our model and the comparing methods under the discriminative setting on VisDial v0.9. Overall, our RvA model outperforms the state-of-the-art methods across all the metrics. Specifically, our RvA model achieves approximately 2 points improvement on R@$k$, and 2$\%$ increase on MRR. In addition, the performance of our model drops significantly without recursive attention (\ie, Rv) or region proposal network (\ie, RPN), which demonstrates their substantial contributions to visual dialog. The similar conclusions can also be drawn on VisDial v1.0 in Table~\ref{tab-1.0-disc}. Furthermore, Table~\ref{tab-0.9-ab} shows a more comprehensive ablation (the component-wise and the feature-wise). It can be seen that by using the proposed recursive attention, any ablative method can be improved regardless of the usage of visual and language representations. Furthermore, our dialog agent could occupy the third place based on the VisDial v1.0 leaderboard\footnote{\url{https://evalai.cloudcv.org/web/challenges/challenge-page/103/leaderboard/298}}, while the team DL-61~\cite{guo2019image} has achieved the best NDCG record 0.5788 in a two-stage fashion.

\input{tab/tab2-VisDial1.0-disc}

We also evaluated the retrieval performance of our model under generative setting on VisDial v0.9. As shown in Table~\ref{tab-0.9-gen}, our approach obtains an approximately 2 points higher R@$k$ compared to the visual co-reference solution model CorefNMN~\cite{kottur2018visual}. In addition, our RvA model outperforms nearly all state-of-the-art methods except CoAtt~\cite{wu2017you}, which is trained using reinforcement learning. 

\input{tab/tab4-VisDial0.9-ablation}
\input{tab/tab3-VisDial0.9-gen}
\subsection{Qualitative Results}
\begin{figure}
    \centering
    \includegraphics[width=0.8\linewidth]{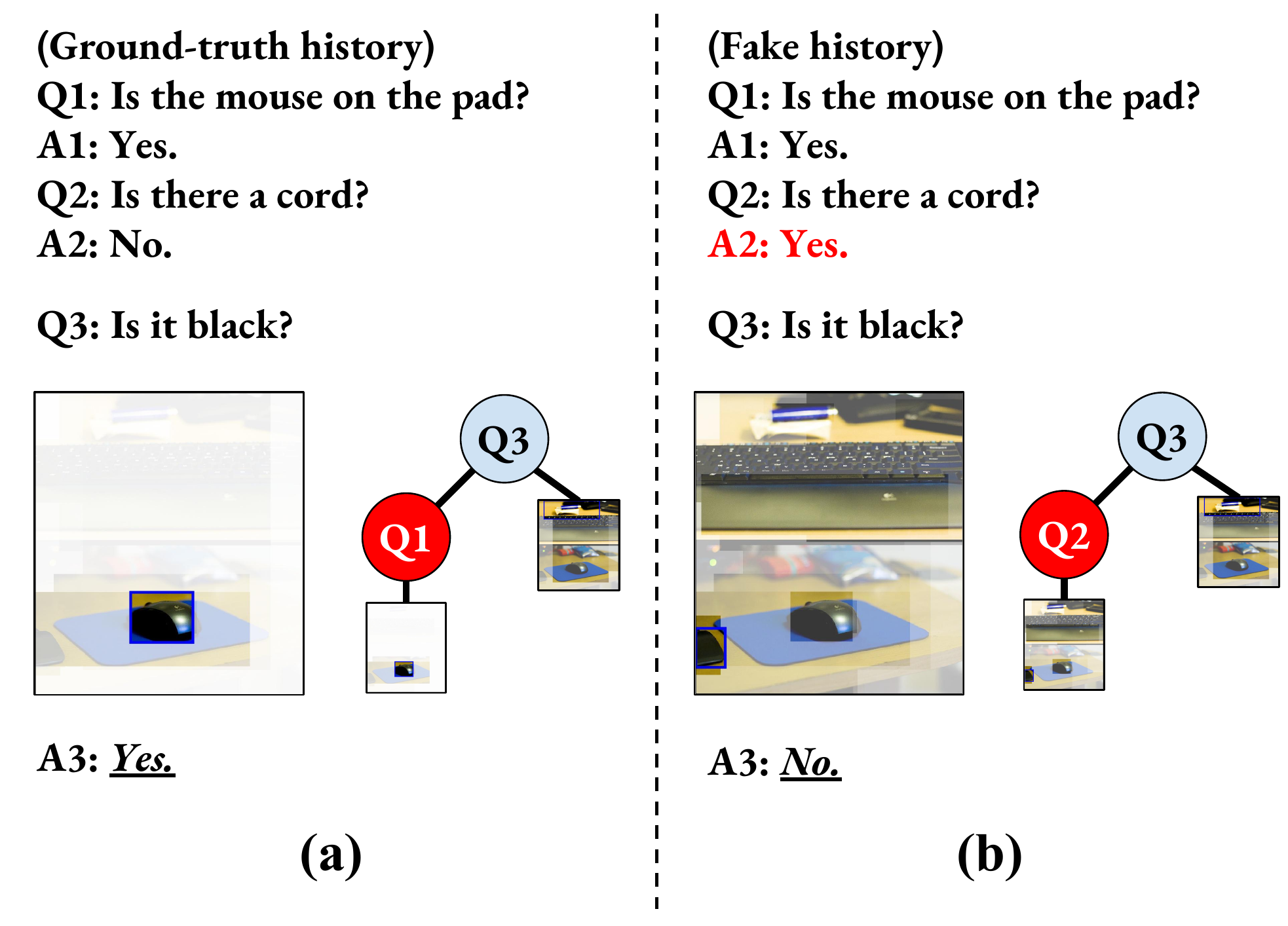}
    \caption{An qualitative example of the history-aware recursion using \module{Pair} module to resolve ``it'' in Q3. (a) represents the recursion obtained by our model using the ground-truth history. Our dialog agent skips the unrelated second history, and pairs the ambiguous question Q3 with the first history. (b) represents the recursion obtained by our model with the fake history, where the second answer ``No'' is replaced with ``Yes''. In this case, our dialog agent pairs the third question with its last history.}
    \label{fig:skip}
    \vspace{-4mm}
\end{figure}

\vspace{-2mm}
The qualitative results shown in Figure~\ref{fig:results} and~\ref{fig:skip} demonstrate the following advantages of our RvA model:

\noindent \textbf{Reasonable Recursions.} Our RvA model achieves reasonable recursions represented by the recursive trees. These recursions can also be regarded as topic-aware dialog clips. Thanks to the reference-aware language feature, our RvA model is able to handle unambiguous sentences with pronouns (\eg, ``Is it sunny outside?'') and ambiguous sentences without pronouns (\eg, ``How many are there?''). Note that it is hard to exhaust all these special cases using a natural language parser.

\noindent \textbf{Reliable Visual Attention.} Our dialog agent successfully focuses on the correct region using recursive visual attention. In contrast, the question-guided visual attention sometimes fails due to the ambiguous question. On the validation set of VisDial v1.0, we observed that: 1) 56\% of question-guided visual attention and 89\% of recursive attention are reasonable for ambiguous questions; 2) 62\% of dialogs require at least one accurate co-reference resolution. Since the recursive visual attention relies heavily on historical visual attention, our dialog agent needs to establish a robust visual attention mechanism. If it were otherwise, the agent would distrust historical visual attention and tend to learn more bias from generic language information, which would hurt the visual dialog system.

\noindent \textbf{History-aware Skipping Pairing.} One may argue that \module{Pair} module can be replaced with referring all the ambiguous questions to their last history (\ie, setting $t_p$ as $t\!-\!1$ in \module{Infer} module) for simplicity. However, our \module{Pair} module is able to \textit{skip} the irrelevant dialog history and produce \textit{history-aware} recursions. As illustrated in Figure~\ref{fig:skip} (a), the dialog agent concludes from the dialog history that ``there is no cord'' in the image. Therefore, the agent skips the second history when pairing the ambiguous question ``Is it black?''. If we replace the second answer ``no'' with ``yes'' to make a fake history (see Figure~\ref{fig:skip} (b)), the third question will be directly paired with its last history. The visual attention and predicted answer are also influenced.

%% file: tab/tab1-VisDial0.9-disc.tex
\begin{table}[t]
\centering
\vspace{1mm}
\scalebox{.82}{
\begin{tabular}{l|ccccc}
\hline 
Model & MRR & R@1 & R@5 & R@10 & Mean \\
\hline
LF \cite{das2017visual}    & 0.5807 & 43.82 & 74.68 & 84.07 & 5.78 \\
HRE \cite{das2017visual}   & 0.5846 & 44.67 & 74.50 & 84.22 & 5.72 \\
HREA  \cite{das2017visual} & 0.5868 & 44.82 & 74.81 & 84.36 & 5.66 \\
MN  \cite{das2017visual}   & 0.5965 & 45.55 & 76.22 & 85.37 & 5.46 \\
HCIAE \cite{lu2017best}    & 0.6222 & 48.48 & 78.75 & 87.59 & 4.81 \\
AMEM \cite{seo2017visual}  & 0.6227 & 48.53 & 78.66 & 87.43 & 4.86 \\
CoAtt \cite{wu2017you}     & 0.6398 & 50.29 & 80.71 & 88.81 & 4.47 \\
CorefNMN \cite{kottur2018visual} & 0.641 & 50.92 & 80.18 & 88.81 & 4.45\\
\hline
RvA w/o RPN   & 0.6436 & 50.40 & 81.36 & 89.59 & 4.22\\
RvA w/o Rv & 0.6551 & 51.81 & 82.35 & 90.24 & 4.07\\
RvA w/o FL & 0.6598 & 52.35 & 82.76 & 90.54 & 3.98\\
RvA            & \bf 0.6634 & \bf 52.71 & \bf 82.97 & \bf 90.73 & \bf 3.93\\
\hline
\end{tabular}
}
\vspace{1mm}
\caption{Retrieval performance of discriminative models on the validation set of VisDial v0.9. RPN, Rv and FL indicate the usage of region proposal network, recursive image attention, and visual feature filter, respectively.}
\vspace{-5mm}
\label{tab-0.9-disc}
\end{table}

%% file: tab/tab2-VisDial1.0-disc.tex
\begin{table}[t]
\centering
\vspace{1mm}
\scalebox{.75}{
\begin{tabular}{l|cccccc}
\hline
Model & MRR & R@1 & R@5 & R@10 & Mean & NDCG \\
\hline
LF \cite{das2017visual}    & 0.5542 & 40.95 & 72.45 & 82.83 & 5.95 & 0.4531 \\
HRE \cite{das2017visual}   & 0.5416 & 39.93 & 70.45 & 81.50 & 6.41 & 0.4546 \\
MN  \cite{das2017visual}   & 0.5549 & 40.98 & 72.30 & 83.30 & 5.92 & 0.4750 \\
CorefNMN$^{\dagger}$ \cite{kottur2018visual} & 0.615 & 47.55 & 78.10 & 88.80 & 4.40 & 0.547\\
\hline
RvA w/o RPN & 0.6060 & 46.25 & 77.88 & 87.83 & 4.65 & 0.5176 \\
RvA w/o Rv & 0.6226 & 47.95 & 79.75 & 89.08 & 4.37 & 0.5319 \\
RvA w/o FL & 0.6294 & 48.68 & 80.18 & 89.03 & 4.31 & 0.5418 \\
RvA & \bf 0.6303 & \bf 49.03 & \bf 80.40 & \bf 89.83 & \bf 4.18 & \bf 0.5559 \\
\hline
\end{tabular}
}
\vspace{1mm}
\caption{Retrieval performance of discriminative models on the test-standard split of VisDial v1.0. ${\dagger}$ indicates that the model uses ResNet-152 features.}
\vspace{-5mm}
\label{tab-1.0-disc}
\end{table}

%% file: tab/tab4-VisDial0.9-ablation.tex
\begin{table}[t]
\centering
\vspace{1mm}
\scalebox{.82}{
\begin{tabular}{ccc|ccccc}
\hline 
 RPN & Bi-LSTM & Rv & MRR & R@1 & R@5 & R@10 & Mean \\
\hline
 & & & 0.6377 & 49.67 & 80.86 & 89.14 & 4.35 \\
 &  & \checkmark & \bf 0.6418 & \bf 50.17 & \bf 81.17	& \bf 89.37	& \bf 4.29\\
\hline
 & \checkmark & & 0.6396 & 49.83 & 81.16 & 89.34	& 4.30 \\
 & \checkmark & \checkmark & \bf 0.6436 & \bf 50.40 & \bf 81.36 & \bf 89.59 & \bf 4.22\\
\hline
 \checkmark &  &  & 0.6534	& 51.78	& 82.28	& 90.21	& 4.09\\
 \checkmark &  & \checkmark & \bf 0.6626	& \bf 52.69	& \bf 82.97	& \bf 90.71	& \bf 3.95\\
\hline
 \checkmark & \checkmark & & 0.6551 & 51.81 & 82.35 & 90.24 & 4.07\\
 \checkmark & \checkmark & \checkmark & \bf 0.6634 & \bf 52.71 & \bf 82.97 & \bf 90.73 & \bf 3.93\\
\hline
\end{tabular}
}
\vspace{0mm}
\caption{Ablations of discriminative models on the validation set of VisDial v0.9. RPN, Bi-LSTM and Rv indicate the usage of region proposal network, bidirectional LSTM, and recursive image attention, respectively.}
\vspace{-3mm}
\label{tab-0.9-ab}
\end{table}

%% file: tab/tab3-VisDial0.9-gen.tex
\begin{table}[t]
\centering
\vspace{1mm}
\scalebox{.82}{
\begin{tabular}{l|ccccc}
\hline 
Model & MRR & R@1 & R@5 & R@10 & Mean \\
\hline
LF \cite{das2017visual}    & 0.5199 & 41.83 & 61.78 & 67.59 & 17.07 \\
HRE  \cite{das2017visual}  & 0.5237 & 42.29 & 62.18 & 67.92 & 17.07 \\
HREA \cite{das2017visual}  & 0.5242 & 42.28 & 62.33 & 68.17 & 16.79 \\
MN  \cite{das2017visual}   & 0.5259 & 42.29 & 62.85 & 68.88 & 17.06 \\
CorefNMN \cite{kottur2018visual} & 0.535 & 43.66 & 63.54 & 69.93 & 15.69\\
HCIAE \cite{lu2017best}    & 0.5386 & 44.06 & 63.55 & 69.24 & 16.01 \\
CoAtt \cite{wu2017you}     & 0.5411 & 44.32 & 63.82 & 69.75 & 16.47 \\
CoAtt$^{\ddagger}$ \cite{wu2017you}     & \bf 0.5578 & \bf 46.10 & \bf 65.69 & 71.74 & 14.43 \\
\hline
RvA w/o RPN    & 0.5417 & 43.75 & 64.21 & 71.85 & 11.18\\
RvA            & 0.5543 & 45.37 & 65.27 & \bf 72.97 & \bf 10.71\\
\hline
\end{tabular}
}
\vspace{1mm}
\caption{Retrieval performance of generative models on the validation set of VisDial v0.9. ${\ddagger}$ indicates that the model is trained using reinforcement learning.}
\vspace{-3mm}
\label{tab-0.9-gen}
\end{table}

%% file: section/5-Conclusion.tex
\section{Conclusions}

In this paper, we formulated the visual co-reference resolution in visual dialog as Recursive Visual Attention (RvA), which consists of three simple neural modules that determine the recursion at run-time. Our dialog agent recursively reviews topic-related history to refine visual attention, and can be end-to-end trained when making discrete decisions of module assembling. Experimental results on the large-scale real-world datasets VisDial v0.9 and v1.0 demonstrate that our proposed model not only achieves state-of-the-art performance, but also obtains explainable recursion and attention maps. Moving forward, we are going to incorporate in-depth language parsing modules into RvA for more accurate recursive decisions.

%% file: section/6-Acknowledgements.tex
\section*{Acknowledgements}
This work was partially supported by National Natural Science Foundation of China (61573363 and 61832017), the Fundamental Research Funds for the Central Universities and the Research Funds of Renmin University of China (15XNLQ01), and  NTU-Alibaba JRI.